\documentclass[conference]{IEEEtran}
\IEEEoverridecommandlockouts
\usepackage{cite}
\usepackage{float}
\usepackage{amsmath,amssymb,amsfonts}
\usepackage{algorithmic}
\usepackage{algorithm}
\usepackage{graphicx}
\usepackage{textcomp}
\usepackage{xcolor}
\usepackage{subcaption}
\usepackage{soul}
\sethlcolor{yellow}
\def\BibTeX{{\rm B\kern-.05em{\sc i\kern-.025em b}\kern-.08em
    T\kern-.1667em\lower.7ex\hbox{E}\kern-.125emX}}

\title{Real-Time Automated donning and doffing detection of PPE based on Yolov4-tiny 
\thanks{\IEEEauthorrefmark{2} These authors contributed equally to this work}
}

\makeatletter
\newcommand{\linebreakand}{%
  \end{@IEEEauthorhalign}
  \hfill\mbox{}\par
  \mbox{}\hfill\begin{@IEEEauthorhalign}
}
\makeatother

\author{
  \IEEEauthorblockN{1\textsuperscript{st} Anusha Verma}
  \IEEEauthorblockA{\textit{CSE Department} \\
    \textit{Wright State University}\\
    Dayton, United States of America  \\
    verma.39@wright.edu}
  \and
  \IEEEauthorblockN{2\textsuperscript{nd} Ghazal Ghajari}
  \IEEEauthorblockA{\textit{CSE Department} \\
    \textit{Wright State University}\\
    Dayton, United States of America  \\
    ghajari.2@wright.edu}
  \and
  \IEEEauthorblockN{3\textsuperscript{rd}  K M Tawsik Jawad}
  \IEEEauthorblockA{\textit{CSE Department} \\
    \textit{Wright State University}\\
    Dayton, United States of America  \\
    jawad.2@wright.edu}
  \linebreakand 
  \IEEEauthorblockN{4\textsuperscript{th} Dr. Hugh P. Salehi }
  \IEEEauthorblockA{\textit{BIHE department} \\
    \textit{Wright State University}\\
    Dayton, United States of America  \\
    Hugh.Salehi@wright.edu}
  \and
  \IEEEauthorblockN{5\textsuperscript{th} Dr. Fathi Amsaad }
  \IEEEauthorblockA{\textit{CSE Department} \\
    \textit{Wright State University}\\
    Dayton, United States of America  \\
    fathi.amsaad@wright.edu}
}
    
\begin{document}

\maketitle

\begin{abstract}
Maintaining patient safety and the safety of healthcare workers (HCWs) in hospitals and clinics highly depends on following the proper protocol for donning and taking off personal protective equipment (PPE). HCWs can benefit from a feedback system during the putting on and removal process because the process is cognitively demanding and errors are common. Centers for Disease Control and Prevention (CDC) provided guidelines for correct PPE use which should be followed. A real time object detection along with a unique sequencing algorithms are used to identify and determine the donning and doffing process in real time. The purpose of this technical research is two-fold: The user gets real time alert to the step they missed in the sequence if they don't follow the proper procedure during donning or doffing. Secondly, the use of tiny machine learning (yolov4-tiny) in embedded system architecture makes it feasible and cost-effective to deploy in different healthcare settings.
\end{abstract}

\begin{IEEEkeywords}
Doffing and Donning, Deep Learning, TinyML, YOLOv4-tiny, Object detection, Real Time, Darknet, Health Safety, PPE \end{IEEEkeywords}

\section{Introduction}
The COVID-19 pandemic and other recent infectious disease outbreaks have highlighted the vital significance that personal protective equipment (PPE) plays in preventing cross-contamination and self-contamination among patients and healthcare personnel\cite{baloh2019healthcare}. For this Centers for Disease Control and Prevention (CDC) has provided protocols to control spread of pathogens \cite{emergency2017emergency}. Studies on the quality of donning and doffing PPE have shown that mistakes are common and adherence to CDC protocols is low \cite{drews2019human}.

\cite{baloh2019healthcare},\cite{salehi2019examining},\cite{chasco2023unfamiliar},\cite{da2020self} Evaluation of PPE adherence in a Polish medical context within the COVID-19 outbreak: An initial case study by \textit{Łukasz Rypicz, Corinne Mowrey, Izabela Witczak, Sandra Furterer, and Hugh Salehi}. Hence, meticulous adherence to PPE protocols is vital to ensure the safety of healthcare personnel\cite{centers2015interim}\cite{jin2020rapid}.

To improve the adherence an automated PPE monitoring systems leveraging real-time object detection can enhance safety by ensuring proper use of PPE. Traditional systems often rely on cloud-based solutions where video frames are sent to a cloud service for analysis. However, these systems face limitations, including dependence on stable network connections, high bandwidth usage, and privacy concerns due to external data processing \cite{balakreshnan2020ppe} \cite{zhou2017security}. An edge-computing approach, where image analysis occurs locally, can mitigate these issues. This method reduces bandwidth requirements, improves system reliability by eliminating reliance on internet connectivity, and preserves worker privacy by processing data locally \cite{sodhro2019mobile}.
\begin{table}[!t]
\setlength{\abovecaptionskip}{0pt}
\renewcommand{\arraystretch}{1.2}
\caption{Objects Images Distribution}
\label{table_example}
\centering
\begin{tabular}{|c|c|}
\hline
\bfseries Object Name & \bfseries No of Images \\
\hline
Coverall & 1157 \\
\hline
Face Shield & 429 \\
\hline
Gloves & 1657\\
\hline
Goggles & 382 \\
\hline
Mask & 1263 \\
\hline
\end{tabular}
\end{table}
\begin{figure*}[!]
\centering
\includegraphics[width=1.0\textwidth,height=0.4\textheight]{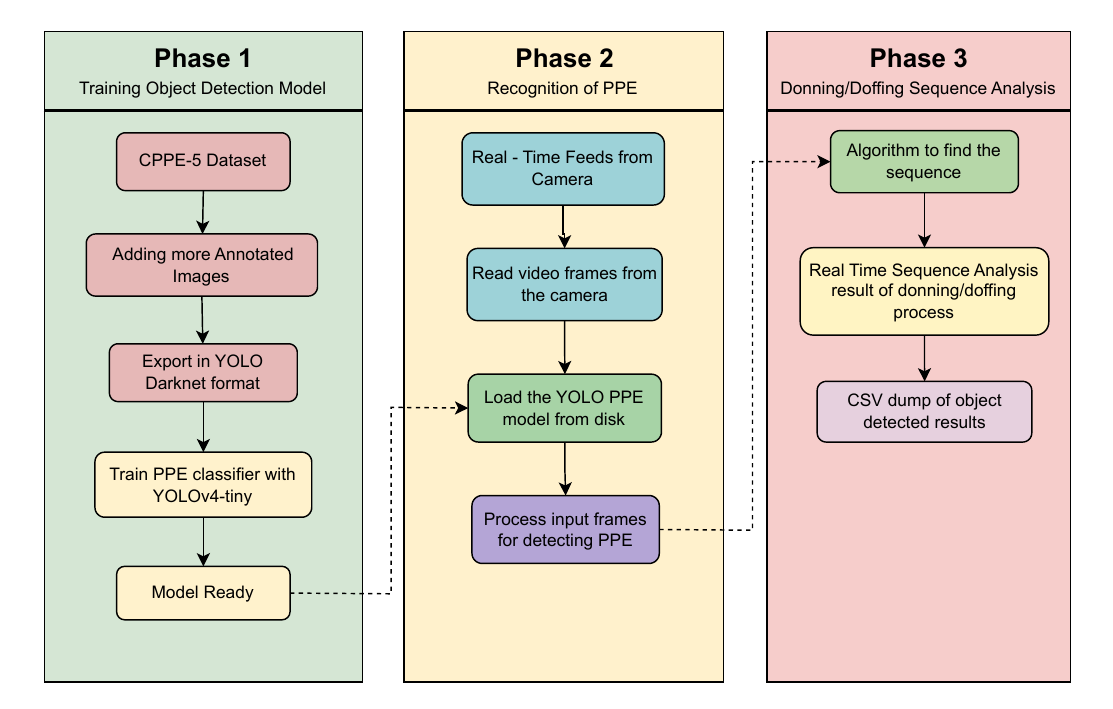}
\caption{ Architecture of the proposed system } \label{fig1}
\end{figure*}

In this work, we provide a real-time PPE detection system that makes use of YOLOv4-tiny deep learning model, specifically designed for healthcare environments. The system developed ensures correct sequence for wearing and removing PPE items such as gloves, masks, and gowns, which is crucial for infection prevention in hospitals.

The following are the primary contributions of this paper:
\begin{enumerate}
    \item Fine-tuning the YOLOv4-tiny model for healthcare settings, evaluating its performance in detecting various PPE items, and validating its real-time capabilities.
    \item Providing real-time feedback on PPE compliance with WHO policy of sequencing PPE equipment, thereby enhancing safety and reducing the risk of infection.
    \item Deploying this system on an embedded platform like the Raspberry Pi, we achieve efficient and low-latency object detection results suitable for real-time sequencing application.
\end{enumerate}

\section{Related Works}
The COVID-19 epidemic has presented the globe with a worldwide health emergency. The necessity of following regulations for the appropriate use, care, and training of personal protective equipment (PPE) was brought home by this situation once more. PPE serves as healthcare workers' last line of defense against the transmission of infections. For personal safety as well as the safety of others, it is crucial to make sure PPE is worn appropriately and consistently\cite{ambigapathy2020should}. HCWs, must follow the CDC guidelines for using  PPE \cite{sauter1992prevention}. Two strategies, based on gown designs, are used by the CDC for general donning and doffing PPE, which involve breaking away and unfastening ties.

Current practices do not provide feedback to HCWs and mistakes cannot be corrected. In a healthcare setting, it is possible for certain PPE to become improperly worn without having monitoring and feedback.\cite{lan2020skin} \cite{schembri2024integrated}.

\section{Architecture}
In this paper, we present a model \cite{bochkovskiy2020yolov4} that analyzes the sequence of donning and doffing of medical PPE Equipment's. We divided our architecture into three phases, each contributing to the overall design and implementation of the architecture.The systems architecture is shown in Fig.\ref{fig1}

\subsection {Phase 1: Dataset and Training Object Detection Model }
The proposed dataset incorporates samples from the CPPE-5 dataset \cite{dagli2023cppe}. Each image in this dataset is annotated with multiple labels and bounding boxes. To enhance the effectiveness and robustness of the data, data augmentation techniques are applied, which also help in preventing model overfitting. The CPPE-5 dataset consists of 1,029 images and 4,698 object annotations across five categories: Coverall, Face Shield, Gloves, Goggles, and Mask, as detailed in Table \ref{table_example}. We used Roboflow tool to annotate the images and do data pre processing and augmentations on the images. The Median height of the images in our dataset is 696x500. During pre processing we resized all images to 600x600. For augmentation we Flip(Horizontal), Rotation(Between -20° and +20°), Bounding Box(Shear: ±10° Horizontal, ±10° Vertical). We finally exported the dataset in YOLO Darknet Format.

\subsection{Phase 2: Object Detection Model for PPE equipment}
The proposed model for detection of PPE Equipments listed in Table \ref{table_example} is based on Yolov4-tiny architecture and consists a total of 38 layers. The architecture incorporates multiple components including convolution layers, routing layers, pooling layers, upsampling layer and yolo layers. These layers work together to extract meaningful features from the input images and perform object detection on the input images. We selected yolov4-tiny so this model can be easily integrated with Raspberry Pi4 and can be easily deployed in remote settings with no wifi to make sure medical staff is still following the doffing and donning process.
YOLO divides the input image into S×S grids of equal sizes where each grid predicts the object located at the center of the grid. Each grid identiﬁes object bounding boxes with an associated conﬁdence score, and boxes with the best conﬁdence scores can be used for object detection. The 4th generation of YOLO (YOLOv4) was released in the early 2020 and is used for PPE gears detection. YOLOv4-Tiny uses CSPDarknet53-Tiny networks for object detection. When finding the sequence of PPE worn, accuracy is not a major effecting factor. 
\begin{figure}[!b]
\centering
\includegraphics[scale=0.72]{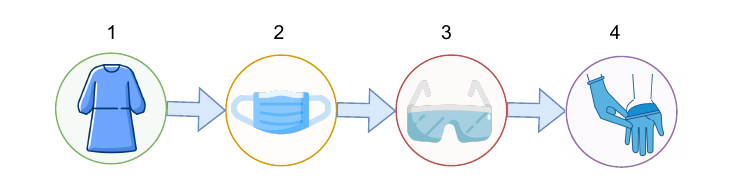}
\caption{Donning(wearing) Sequence Gown, Mask, Googles/Face shield, Gloves}
\label{fig2}
\end{figure}
\subsection{Phase 3: Doffing/Donning Sequence Analysis}
The risk of contracting respiratory and other healthcare-associated diseases is considerable for healthcare workers (HCWs),\cite{macintyre2014quantifying} and part of this risk is due to self-contamination during the removal (or "doffing") of contaminated personal protective equipment (PPE).\cite{mclaws2016highly}\cite{tomas2015contamination}\cite{zellmer2015variation}.Recently we witnessed the Pandemic Covid-19, in which the concern was for the HCW to not come in contact with the virus during removing(doffing) the PPE. We developed a checklist based on hospital protocols for donning and doffing PPE (Appendices 1a and 1b), tailored to the PPE items listed in Table \ref{table_example}. This sequencing protocol complies with the World Health Organization (WHO) \cite{world2020rational} regarding PPE use in the context of COVID-19. Artificial intelligence, particularly computer vision techniques such as object detection in medical imaging, has seen widespread adoption \cite{ragab2024comprehensive}. CNN models are commonly used for object detection. There are different kinds of detectors: one-stage \cite{liu2016ssd} and two-stage detectors\cite{redmon2016you, jiang2020real}\cite{diwan2023object}. When it comes to accuracy of the model, two stage detectors are better, however if the main objective is faster inference than two-stage detector is better.

\begin{algorithm}
\caption{Real-time Sequence Detection Algorithm}
\begin{algorithmic}[1]
\STATE Procedure RealtimeSequenceDetection
\FOR{$i = 1$ to $n$}
    \IF{PPE meets $\text{thFrequency}$ and $\text{thConfidence}$ criteria}
        \STATE Mark the task as done and record the time frame
        \STATE \textbf{break} out of loop
    \ELSIF{Mask meets $\text{thFrequency}$ and $\text{thConfidence}$ criteria}
        \STATE Mark the task as done and record the time frame
        \STATE \textbf{break} out of loop
    \ELSIF{Gloves meet $\text{thFrequency}$ and $\text{thConfidence}$ criteria}
        \STATE Mark the task as done and record the time frame
        \STATE \textbf{break} out of loop
    \ELSIF{FaceShield meets $\text{thFrequency}$ and $\text{thConfidence}$ criteria}
        \STATE Mark the task as done and record the time frame
        \STATE \textbf{break} out of loop
    \ELSIF{Glasses meet $\text{thFrequency}$ and $\text{thConfidence}$ criteria}
        \STATE Mark the task as done and record the time frame
        \STATE \textbf{break} out of loop
    \ENDIF
\ENDFOR
\STATE Check if donning or doffing sequence is followed
\STATE \textbf{return} Result
\STATE End Procedure
\end{algorithmic}
\end{algorithm}

\begin{figure}[!t]
\centering
\includegraphics[scale=0.72]{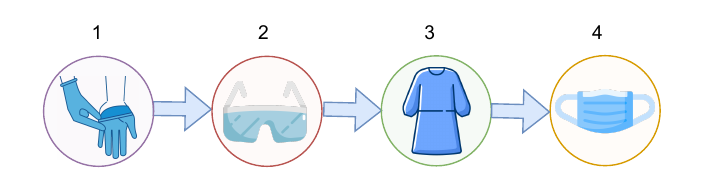}
\caption{Doffing(removing) Sequence Gloves, Googles/Face shield, Gown, Mask- Unfastening Ties Strategy}
\label{fig3}
\end{figure}

Without getting into the intricacies of the PPE donning and doffing procedure, some study has been done on the use of AI for face mask identification\cite{teboulbi2021real} and other PPE equipment detection\cite{kumar2020proposed}. This study aims to develop a model that verifies the correct donning and doffing procedures. Currently, the process relies on monitoring by trained educators. This project advances existing research by creating an innovative algorithm that could (1) ensure healthcare professionals (HCPs) follow PPE guidelines for donning and doffing, and (2) provide an innovative tool to support HCP educators in creating simulation environments that surpass traditional teaching methods.
For determining the sequence of doffing and donning of PPE gears worn from results of YOLOv4-tiny in real time we have come up with an algorithm. The detection of any object using the object detection model is highly associated with mAp obtained during the training of model for each class. We calculate the threshold for each class separately and once the threshold values are accomplished we mark the PPE item as worn. The sequence for which we will be checking is shown in Fig.\ref{fig2} and Fig.\ref{fig3}. 
The Real-time Sequence Detection algorithm presented in this study serves as a pivotal component in our research endeavor, aimed at enhancing the efficiency and accuracy of personal protective equipment (PPE) monitoring systems. The algorithm below shows the pseudo code for the sequencing algorithm. thFrequency is the Threshold Frequency and thConfidence is the Threshold Confidence value calculated from the mAP value for each class. The correct calculation of threshold values is very important to successfully determine if the correct sequence or pattern is followed or not. Overall, the Real-time Sequence Detection algorithm represents a significant advancement in monitoring PPE usage patterns, with implications for improving safety protocols, enhancing data-driven decision-making, and ultimately mitigating risks associated with infectious diseases or workplace hazards.

\section{Experimental Results}
We have utilized a open-source CNN framework written in C and CUDA, Darknet \cite{redmon2013darknet} We imported pre-trained weights from AlexeyAB's Darknet for the purpose of training the model. More specifically, the yolov4.conv.137 weights were used to train the YOLOv4-tiny model. The filter size for the yolov4-tiny model was calculated using the equation $(\text{class} + 5) \times 3$. This implementation was done using an online NVIDIA Tesla A100 GPU that is accessible through Google Colab.

\begin{figure}[!t]
\centering
\includegraphics[scale=0.4]{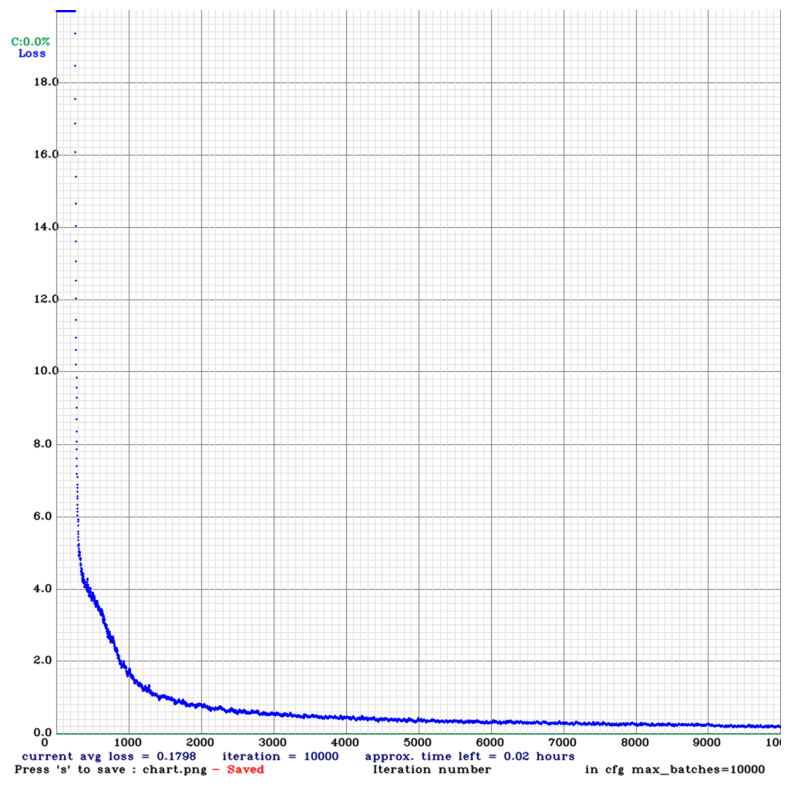}
\caption{Loss Graph for the Yolov4-tiny model}
\label{fig4}
\end{figure}

\begin{figure}[!t]
\centering
\includegraphics[scale=0.1]{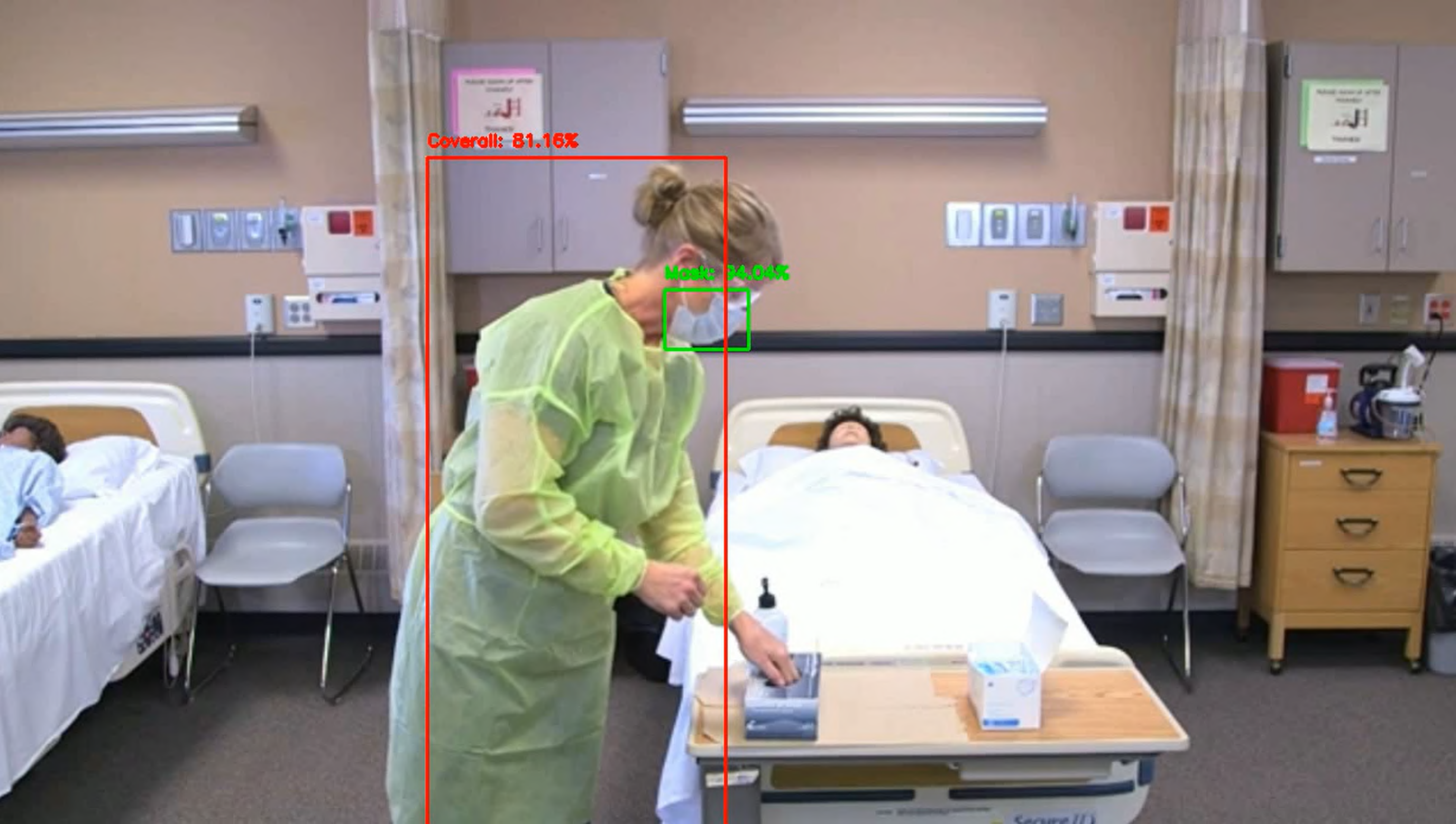}\\
(b)\\
\includegraphics[scale=0.1]{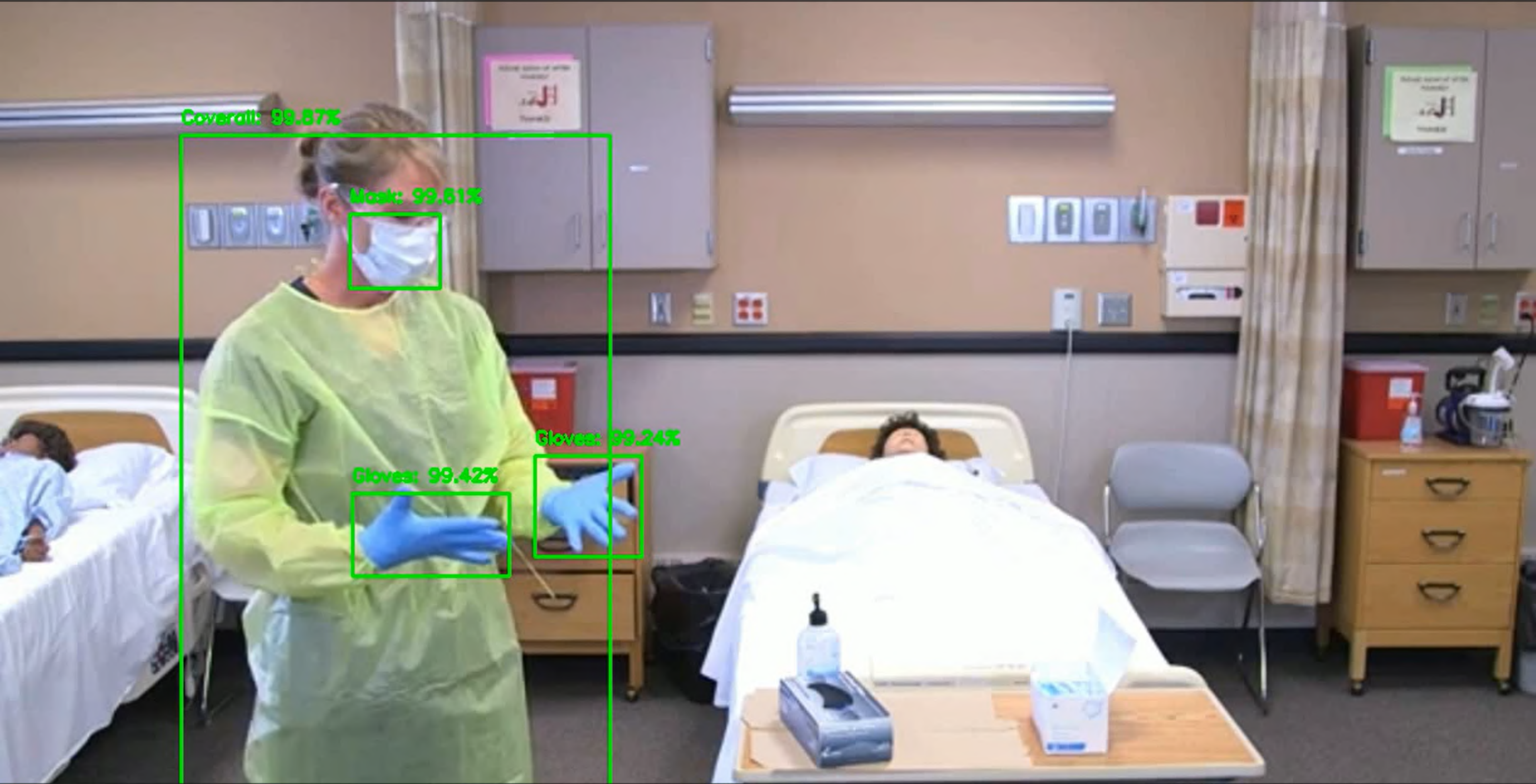}\\
(c)\\
\caption{output video snapshots.}
\label{fig5}
\end{figure}

After training for 45 minutes, the Tiny YOLOv4 models attained 10,000 iterations. Analysing the graph in Fig.\ref{fig4}, we can conclude the loss factor for Tiny YOLOv4 was 0.1798 at the training step 7000. A YouTube video with the proper donning sequence was chosen to validate the trained Tiny YOLOv4 model's prediction results and check for our sequencing algorithm. The results in Fig.\ref{fig5} shows the snapshots taken after detection was performed on the video. we have set confidence threshold for each class. In Fig.\ref{fig5}(b) PPE shows less confidence than the threshold confidence.

\section {Conclusion}
In this study, a PPE wearing automated real time detection and task sequencing model based on YOLOv4-tiny is proposed for medical professionals. This model uses deep learning techniques to intelligently identify between various PPE products. Our study's findings suggest that this model can be utilized to consistently and effectively track medical staff members' PPE donning and doffing situations, lessen the possibility of injury from human subjectivity throughout the management process, and conserve medical resources.

Moreover, Our research demonstrates the feasibility of integrating our object detection model with a Raspberry Pi 4 for real-time action in remote areas. We have demonstrated that it is feasible to create autonomous and responsive systems in resource-constrained contexts by utilizing the Raspberry Pi's low power consumption, small form factor, and processing capacity in conjunction with our optimized custom model.

\bibliographystyle{IEEEtran}
\bibliography{IEEEabrv,Bibliography}

\end{document}